\documentclass[conference]{IEEEtran}
\IEEEoverridecommandlockouts
\usepackage{cite}
\usepackage{amsmath,amssymb,amsfonts}
\usepackage{algorithmic}
\usepackage{graphicx}
\usepackage{subfigure}
\usepackage{textcomp}
\usepackage{xcolor}

\usepackage{multirow}
\usepackage{hyperref}
\usepackage{mathrsfs}
\usepackage{CJKutf8}

\def\BibTeX{{\rm B\kern-.05em{\sc i\kern-.025em b}\kern-.08em
    T\kern-.1667em\lower.7ex\hbox{E}\kern-.125emX}}
\begin{document}

\title{Jointly Learning Span Extraction and Sequence Labeling for Information Extraction \\from Business Documents
}

\author{\IEEEauthorblockN{Nguyen Hong Son\IEEEauthorrefmark{2}, Hieu M. Vu\IEEEauthorrefmark{2}, Tuan-Anh D. Nguyen\IEEEauthorrefmark{2}, Minh-Tien Nguyen\IEEEauthorrefmark{2}\IEEEauthorrefmark{3}\IEEEauthorrefmark{1}\thanks{*Corresponding Author.}}


 \IEEEauthorblockA{\IEEEauthorrefmark{2}CINNAMON AI, 10th floor, Geleximco building, 36 Hoang Cau, Dong Da, Hanoi, Vietnam.\\ 
 Email: \{levi, ian, tadashi, ryan.nguyen\}@cinnamon.is}
  
 \IEEEauthorblockA{\IEEEauthorrefmark{3}Hung Yen University of Technology and Education, Hung Yen, Vietnam.\\
 Email: tiennm@utehy.edu.vn} 
 
}


\IEEEpubid{\begin{minipage}{\textwidth}\ \\[12pt] \centering
  1551-3203 \copyright 2022 IEEE. Personal use of this material is permitted.  Permission from IEEE must be obtained for all other uses, in any current or future media, including reprinting/republishing this material for advertising or promotional purposes, creating new collective works, for resale or redistribution to servers or lists, or reuse of any copyrighted component of this work in other works.
\end{minipage}}

\maketitle

\begin{abstract}
This paper introduces a new information extraction model for business documents. Different from prior studies which only base on span extraction or sequence labeling, the model takes into account advantage of both span extraction and sequence labeling. The combination allows the model to deal with long documents with sparse information (the small amount of extracted information). The model is trained end-to-end to jointly optimize the two tasks in a unified manner. Experimental results on four business datasets in English and Japanese show that the model achieves promising results and is significantly faster than the normal span-based extraction method. The code is also available.\footnote{https://bit.ly/3iYztCL. It will be available on github.}
\end{abstract}


\section{Introduction}
Information extraction (IE) is an important task of natural language processing (NLP), in which IE models extract information from a given document \cite{APM-ACL-IJCNLP-15,LBSKD-NER-NACCAL-16,Ram-Few-shot-span-IE-21,Nguyen-Recur-Span-DIL-21,Li-Span-NER-21}. The extraction task can be done by using machine learning \cite{APM-ACL-IJCNLP-15,LBSKD-NER-NACCAL-16,Zhang-rapid-adaptation-BERT-20,Karamanolakis-TXtract-ACL-20}, in which IE models are trained by using hand-crafted features \cite{APM-ACL-IJCNLP-15} or hidden features automatically learned from data \cite{LBSKD-NER-NACCAL-16,Karamanolakis-TXtract-ACL-20}. Recently, pre-trained language models (PreLMs), e.g. BERT \cite{DCLT-NAACL-19} or LayoutLM \cite{Xu-LayoutLM-20} have been adapted to IE of business documents with promising results \cite{Zhang-rapid-adaptation-BERT-20,Nguyen-Recur-Span-DIL-21}. In practice, extracted information can be used for the digital transformation \cite{Herbert-dig-tran-17,Karamanolakis-TXtract-ACL-20} of companies who want to digitally manage their data for efficient office operation \cite{Zhang-rapid-adaptation-BERT-20}.

\begin{figure}[!h]
    \centering
  \includegraphics[scale=0.5]{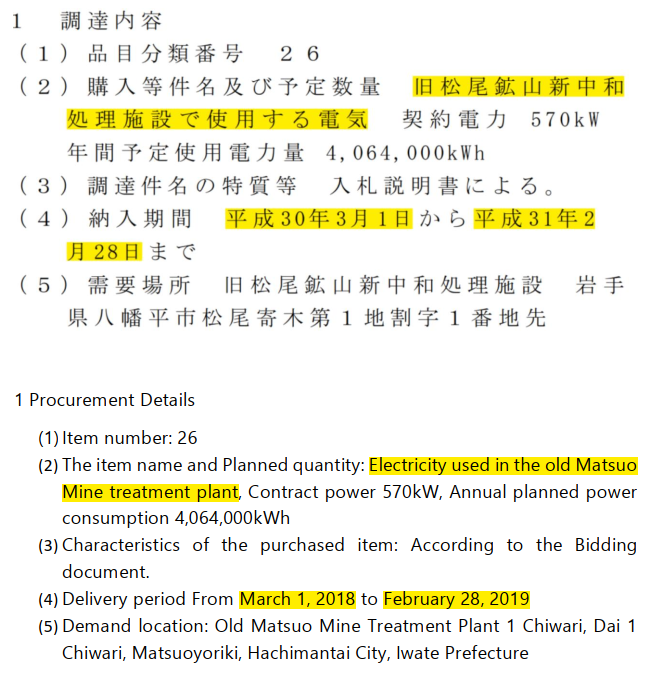}
    \caption{A paragraph from a bidding document. The upper part is the original Japanese text and the bellow part is the translation. Yellow spans are information that need to be extracted. The first yellow line in the clause 2 is the bid subject. The second yellow line in the clause 4 is the procurement start and end date. This data is a contract document in the Bidding dataset \cite{Nguyen-IE-IJCNN-21}. Documents were taken from a Japanese public website.}\vspace{-0.3cm}
\label{fig:bidding_doc_example}
\end{figure}

We focus on information extraction from business documents, which can help companies to reduce the cost of data management and transformation \cite{Zhang-rapid-adaptation-BERT-20,NLL-EAAI-21}. Compared to traditional data types, e.g. news (CoNLL 2003 \cite{SKM-CONLL-03}), business documents are quite long (Table \ref{tab:data}), complicated, and specific in writing styles. For example, a bidding document is usually long, i.e. 30-50 pages, including a lot of information for bidding. Figure \ref{fig:bidding_doc_example} shows five claims in a bidding article, important information which needs to be extracted is highlighted by yellow. For instance, given two queries: \texttt{``the bid subject"} and \texttt{``the procurement period"}, IE models have to extract relevant spans. The span of \texttt{``the bid subject"} is in the second clause and the spans of \texttt{``the procurement period"} (procurement start and dates) are in the in the fourth clause.

\IEEEpubidadjcol




There are two main methods for IE: span extraction formulated as machine reading comprehension \cite{Ram-Few-shot-span-IE-21,Nguyen-Recur-Span-DIL-21,Li-Span-NER-21} and sequence labeling, also called named entity recognition (NER) \cite{LBSKD-NER-NACCAL-16,Li-MRC-NER-ACL-20}. 
The two methods have achieved promising results for extracting information \cite{Huang-BiLSTM-CRF-NER-15,Katiyar-Nested-NER-NAACL-18,DCLT-NAACL-19,Zhang-rapid-adaptation-BERT-20,Ram-Few-shot-span-IE-21}, yet long business documents with very sparse information challenge the normal IE techniques. On the one hand, the span extraction approach is good at dealing with long documents with sparse information \cite{DCLT-NAACL-19,Zhang-rapid-adaptation-BERT-20} because it directly extracts answer spans based on their start and end positions. However, it could not completely handle the problem of multiple span extraction, which is popular in actual business cases \cite{Li-MRC-NER-ACL-20,Nguyen-Recur-Span-DIL-21}. For example, the bidding dataset in Table \ref{tab:data} has 27.25\% of multiple answer samples. In addition, the speed of normal span extraction methods (e.g. BERT-QA) creates an obstacle in practical cases because it sequentially encodes each question-context pair. With many questions (tags) and long context documents (Table \ref{tab:data}), the training and inference time is unpractical ($m$ times to encode $m$ query-document pairs). On the other hand, sequence labeling can handle multiple span extraction \cite{Huang-BiLSTM-CRF-NER-15,Katiyar-Nested-NER-NAACL-18}; however, its quality tends to be reduced due to sparse information. We observed that the percentage of answer tokens in business documents is tiny (i.e. 0.62\% of bidding). Hence, training a high-quality sequence labeling model for such data is a non-trivial task \cite{LBSKD-NER-NACCAL-16,Li-MRC-NER-ACL-20}.


This paper introduces a new IE model, which takes the advantage of the span-based extraction and sequence labeling (NER) methods for business documents. The combination allows the model to extract multiple spans from long business documents with sparse information. To speed up the training process, we design a new simple but effective method for encoding input queries (short fields, e.g. bidding date) and long context documents. Instead of putting each field-document (context) pair into the encoder, the model considers the document and input fields independently. This allows the model to encode short fields and the context in parallel by using multiple general attention. The model is trained end-to-end to jointly optimize the two tasks. Final outputs are created by using an aggregation algorithm based on span prediction and sequence labeling. This paper makes two main contributions:
\begin{itemize}
    \item It introduces a practical IE model, which combines span extraction and sequence labeling in a unified model. The speed of the training and inference process is also significantly improved by using two-channel information encoding. The model achieves high accuracy of information extraction from business documents with limited data, which is applicable for business use.
    
    \item It validates the model on four business datasets in English and Japanese. The analyses also show the behavior of the model in several aspects: F-scores, training and inference time, convergence, and the coverage of multiple spans.
\end{itemize}

\section{Related Work}
\subsubsection{Named entity recognition} Named entity recognition (NER) is a widely studied task in NLP. NER models generally fall into two main categories: tag-based approaches and span-based approaches. For tag-based approaches, entity extraction is treated as a sequence labeling problem \cite{Liu-NER-AAAI-18}, with common architectures including: bidirectional  LSTM-CRF \cite{LBSKD-NER-NACCAL-16,JWA-NAACL-18}, CNN \cite{Collobert-NLP-JML-11} and seq2seq \cite{Liu-NER-AAAI-18}. With the advances of pre-training methods, models using contextual word representation such as Flair \cite{Akbik-Flair-COLING-18} or BERT \cite{DCLT-NAACL-19} further enhanced the performance of NER, yielding state-of-the-art results. However, tag-based models still suffer from the nested NER problem, where overlapping entities can not be resolved efficiently \cite{btg1023}. Also, common tag-based models can only operate on short sequences (i.e. sentences), so its adaptation to long business documents is still an open question. More recently, span-based approaches have attracted interest due to their generalization ability in both flat and nest NER tasks \cite{Zheng-Boundary-Nested-NER-EMNLP-19,Fu-Nested-NER-TreeCRF-AAAI-21}.



\subsubsection{Span extraction} The span extraction was originally derived from the machine reading comprehension (MRC) problem that extracts answer spans from a context passage given an input question \cite{Rajpurkar-SQuAD-EMNLP-16}. The MRC problem can be formulated as a classification task that predicts the start and end positions of answer spans.  Recent studies have been adapted MRC for NER \cite{Li-MRC-NER-ACL-20,Li-Span-NER-21}. For example, \cite{Li-MRC-NER-ACL-20}  introduced a unified MRC model for NER. Instead of using sequence labeling, the model formalizes the extraction as MRC which can deal with flat and nested entities. \cite{Li-Span-NER-21} presented a span-based model for jointly extracting overlapped and discontinuous entities. \cite{aly-etal-2021-leveraging} proposed a cross-attention model between context and entities description to improve NER performance in zero-shot scenarios. The success of transformers, i.e. BERT \cite{DCLT-NAACL-19} substantially facilitates the span extraction progress. Several studies have been utilized BERT for IE from business documents \cite{Zhang-rapid-adaptation-BERT-20,NLL-EAAI-21} by using the Question Answering (QA) formulation.

\subsubsection{Our uniqueness}
Our study is inspired by the work of \cite{Li-MRC-NER-ACL-20} who presented a unified framework for NER. However, in contrast to \cite{Li-MRC-NER-ACL-20} which has to create long questions based on entities, we instead design a new encoding method that separately encodes short fields and long documents. The encoding method allows speeding up the training process. We share the idea of using MRC for IE on business documents \cite{Zhang-rapid-adaptation-BERT-20,NLL-EAAI-21}; however, our model can tackle the extraction of multiple spans which was not mentioned in previous works \cite{Zhang-rapid-adaptation-BERT-20,NLL-EAAI-21}. We also share the idea of recurrent decoding with \cite{Nguyen-Recur-Span-DIL-21} for multiple span extraction of business documents, but our method can further handle long context.

\section{The Proposed Model}\label{sec:model}

\begin{figure*}[!h]
	\centering
	\includegraphics[width=0.9\textwidth]{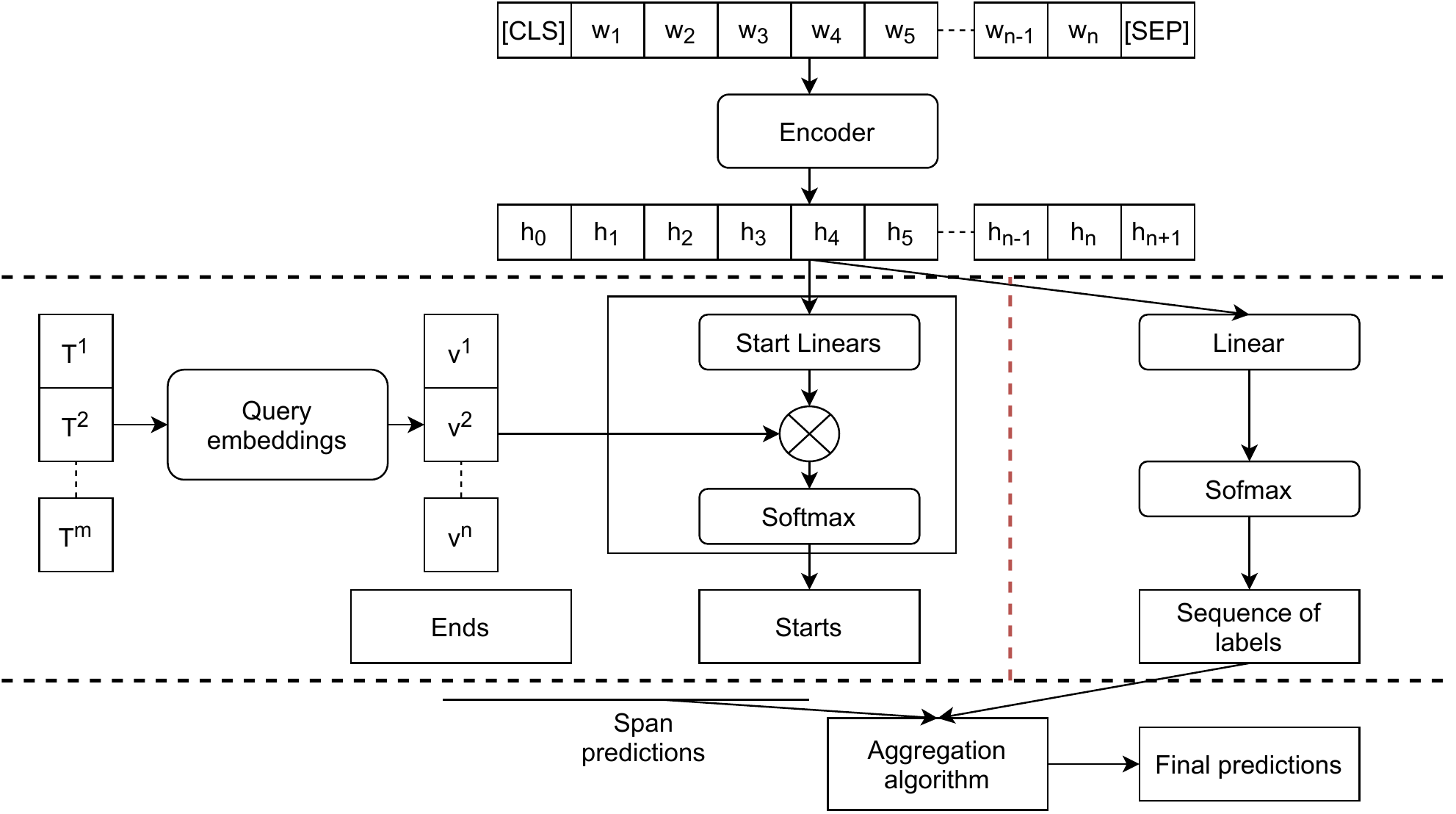}
	\caption{The proposed model which jointly learns span extraction and sequence labeling for information extraction.}\label{fig:model}
\end{figure*}

\subsection{Problem formulation}
This work focuses on the extraction of predefined information from business documents, e.g. bidding. Let $D =\{w_{1}, w_{2},...,w_{n}\}$ is the target document with $n$ tokens, $T= [T^{1}, T^{2},...,T^{m}]$ is the set of fields (tags) which are unchanged in training and inference, and each field $T^{i} = \{ t^{i}_{1}, ..., t^{i}_k \}$ is a set of $k$ tokens ($k << n$). IE models extract the most appropriate segment of text by learning the mapping function $f(D,T|\Theta) \label{pro_for}$, in which $\Theta$ can be learned by using span extraction \cite{DCLT-NAACL-19} or sequence labeling \cite{LBSKD-NER-NACCAL-16,Li-MRC-NER-ACL-20}. For span extraction, the fields in $T$ are input queries combined with the context document $D$ for encoding. For sequence labeling, the fields in $T$ are used for data annotation by using the BIO format. We use "queries" and "fields" as the same meaning. The model receives a short tag as a question (QA formulation) or label (NER) and then pulls out a relevant span from the document. For example, given an input tag of \texttt{"contract date"}, the model extracts the date of the contract.



\subsubsection{Model description}
Based on the formulation, we design a model for jointly learning span extraction and sequence labeling in Fig. \ref{fig:model}. The model encodes each input context document $D$ by using a strong PreLM (e.g. BERT) to obtain the contextualized vectors. The span extraction part (on the left, called the SpanIE part) encodes short fields by using query embedding to output query vectors. These vectors are fused with the contextualized vectors from the encoder by using attention for dealing with long documents and sparse information. Meanwhile, the contextualized vectors are also used for sequence labeling (on the right, called the NER part) for extracting multiple spans. Final candidate spans are extracted by using an aggregation algorithm. The next sections describe the detailed information of our model.

\subsection{Document encoding}
Document encoding is the first step that map input documents to vector representation for learning. To do that, we used pre-trained language models (PreLMs), e.g. BERT \cite{DCLT-NAACL-19} for encoding input context documents. This is because BERT has shown its efficiency to represent the context of tokens, which is critical for IE \cite{DCLT-NAACL-19,Zhang-rapid-adaptation-BERT-20,NLL-EAAI-21}. Given an input document $D$, all $n$ tokens of $D$ were concatenated to form a single sequence. It was fed into the encoder of BERT \cite{DCLT-NAACL-19} to obtain the hidden representation of each token $h_{j} \in  \mathbf{R}^{c}$. Using BERT is to deal with data limitation in business cases \cite{Zhang-rapid-adaptation-BERT-20,Nguyen-Recur-Span-DIL-21}. The output of document encoding is the sequence of contextualized encoded vectors, each corresponding to an input token:
\begin{equation}
    \boldsymbol{H} = \{h_{0}, h_{1},...., h_{n}\}
\end{equation}
where $h_{i}\in \mathbb{R}^{d}$ is the encoded vector of the $i^{th}$ token; $n$ is the number of input tokens. The contextualized vectors from BERT are the inputs for span extraction and sequence labeling.

\subsection{Span extraction}
We observed that the original span extraction based on the question answering formulation by using BERT (BERT-QA) has achieved promising results \cite{DCLT-NAACL-19}, yet it needs a long time for training and inference (Table \ref{tab:training-inference}). It comes from the nature of BERT-QA that sequentially takes each pair of query-context document for encoding. In this section, we introduce a new design, that not only takes advantage of span prediction which obtains high accuracy for extracting single spans but also speeds up the training and inference of our model. This part includes query embedding and attention.

\subsubsection{Query embedding}
As mentioned, the extraction task was formulated as question answering (QA), that the model receives a question and pulls out the relevant answer from the context document. However, we argue that the combination of a short field and a long document (see Table \ref{tab:data}) as the normal span extraction method is not necessary for our task because the information of a short field (a few words, e.g. contract value) might be saturated by the information of a long document. So, the representation from BERT tends to be biased to the document. It makes a challenge in capturing the meaning of a short field. To tackle this issue, we separately encoded all fields using a simple embedding layer to create an embedding matrix $V \in \mathbf{R}^{m \times d}$, with $m$ is the number of fields and $d$ is the embedding size. Each row $v^i$ of $V$ represents the embedding of the corresponding field $T^i$ and is learnt during training. Hence the query embedding process for a field $T^i$ is simply extracting its corresponding row in $V$. This method has two advantages. First, the information of short queries is kept in a different channel. It allows the model to reduce the bias of long documents in the training process. Second, it allows the model to encode all short fields at the same time, which reduces the complexity of the encoding process from $O(m)$ to $O(1)$. This makes the uniqueness of our model compared to the original BERT-QA \cite{DCLT-NAACL-19}. For embedding, we randomly initialized vectors since the results of using BERT are similar.

\subsubsection{Multiple general attention}
The query fields (tags) are separately encoded by the embedding layer, so the encoder itself can not capture the latent relationship between the fields and the context effectively. The information of fields can only affect the encoder via the gradient update during back-propagation, which adjusts parameters for better encoding the context. We, therefore, adapted the general attention \cite{Luong-Att-EMNLP-15} to encode the latent relationship of the fields and the context. Let each tag embedding is the target vector and each token hidden state of the context is the source vector, the general attention estimates the correlation between the target and the source vectors.

In this step, to extract the start positions (represented by the \texttt{Starts} block in \ref{fig:model}), the token embeddings first go through a series of fully-connected layers (the \texttt{Start Linears} block in \ref{fig:model}). The output is then used in a matrix multiplication with the query embeddings to get the attention score. Finally, the softmax function is applied, giving the start positions of the spans. A similar process with a different sets of weights are used to caculated the end positions (the \texttt{End} block in \ref{fig:model})

It is possible to use one attention layer for all fields. However, a single layer can not completely represent the complex signal from all fields. Also, all fields are trained in parallel which enables the model to use multiple attention. For each tag, we used two attention layers to independently updates the weight matrix of the start and end positions to retain the signal of the field. Let $v^{i} \in \mathbf{R}^{d} $ be the embedding vector of the field $T^i$, $h_j$ is the contextualized vector of each token $j^{th}$ of $D$, and $p^{i}_{sj}$ is the probability of the start position $j^{th}$ of the answer corresponding to the field $T^i$. $p^{i}_{sj}$ is calculated as follows.
\begin{equation}
    p^{i}_{sj} = \frac{\exp (score(v^{i\top}, h_{j}))}{\sum_{j'} \exp (score(v^{i}, h_{j'})} \\
    =\frac{\exp (v^{i\top}W^{i}_{s} h_{j}))}{\sum_{j'} \exp (v^{i\top}W^{i}_{s} h_{j'})}
\end{equation}
with $W^{i}_{s} \in \mathbf{R}^{d \times c}$ is the weight matrix of the start attention layer. Similarly, the end position is computed as follows.
\begin{equation}
    p^{i}_{ej} = \frac{\exp (v^{i\top}W^{i}_{e} h_{j}))}{\sum_{j'} \exp (v^{i\top}W^{i}_{e} h_{j'})}
\end{equation}

The start and end positions of the field $T^i$ were calculated as $s^{i}=\textrm{\textit{argmax}}_{j}(p^{i}_{sj})$ and $e^{i}=\textrm{\textit{argmax}}_{j}(p^{i}_{ej})$ for pulling the relevant text span of $T^i$ from $D$.

\subsection{Sequence labelling}
As mentioned, the span extraction method can not completely handle multi-span extraction. To address this problem, we empower our model by taking the advantage of sequence labeling (NER) for extracting multiple answer spans. Our motivation comes from the fact that business documents are very long (Table \ref{tab:data}) which brings a challenge to the encoder of BERT. In fact, it can not encode the sequences longer than 512 tokens. Therefore, it is really challenging to encode the whole document. By applying a stride window, we can split the document into multiple sub-sequences, each has an overlap with the previous and next sub-sequences.  In this configuration, a token can belong to different sub-sequences, and its encoded vector does not remain constant. To get the final encoded vector of each token, we calculated the average vector corresponding to the token over all sub-sequences that the token was included.
\begin{equation}
    h_j = \frac{\sum_{S} h_{j'}}{|S |}
\end{equation}
with $S$ is the set of sub-sequences that contains the considering token $j^{th}$ and $h_{j'}$ is the encoded vector of the token in each sub-sequences.

In order to predict the label of each token, its encoded vector was simply passed into a stack that includes a dense layer followed by a softmax layer. Let $n_{e}$ be the number of labels in the BIO format, the probability of the label $i^{th}$ for the corresponding token $j^{th}$ is calculated as follows.
\begin{equation}
    p_{j}^{i} = \frac{exp(Wh_{j})}{\sum_{j'}exp(Wh_{j'})}
\end{equation}
with $W \in \mathbb{R}^{c \times n_{e}}$ is the weight matrix of NER. The predicted outputs of the span IE and NER were aggregated to produce the final result.


\subsection{The aggregation algorithm}
The span-IE and NER parts produce two different outputs. The span IE prediction is very precise for the first answer span while the NER can predict multiple spans. To combine the output of the two methods, we simply set a higher priority to the span IE part. In particular, for each entity type, if there is any entity from the NER output that is overlapped with the span-IE output, we use the span output. The final output is the combination of span-IE prediction and all predicted entities which are not overlapped with the span-IE output.

\subsection{Training and inference}
For training, both span extraction and sequence labeling were trained jointly by using the cross-entropy loss. Suppose the span extraction was trained by the loss $\mathcal{L}_{span}$, the sequence labeling has the loss as $\mathcal{L}_{NER}$, we tried two methods for combining these two losses. The first method is simply sum up the two losses as follows.
\begin{equation}\label{eq:simple-loss}
    \mathcal{L} = \mathcal{L}_{span} + \mathcal{L}_{NER}
\end{equation}
Another way to compute the final loss is using linear combination with learnable factor:
\begin{equation}
    \mathcal{L} = \alpha \mathcal{L}_{span} + (1- \alpha)\mathcal{L}_{NER}
\end{equation}
with $\alpha$ is the learnable factor that is learned during training. However, our experimental results showed that there is no significant difference between these two methods in term of F-scores. Therefore, we used the simple version of the combination loss in Eq. \eqref{eq:simple-loss} for training the model.

For inference, given the hyper-parameter $\Theta$, an input document $D$, and a field $T^i \in T$, the model predicts the start and end positions of the field $T^i$ by using span extraction and a set of sequence labels by using sequence labeling. Final extracted spans are the combination of the two predicted outputs.

\section{Settings and Evaluation Metrics}
\subsection{Data}
We confirm the efficiency of the proposed model on internal and benchmark datasets as follows.
\subsubsection{Internal data}
We used a Japanese Permit \textit{ License} dataset for evaluation. IE models are required to extract 12 information types from each document. There are 328 training documents and 104 documents for testing.

\begin{table}[!h]
\centering
\setlength{\tabcolsep}{2pt}
\caption{Four business data sets. \textit{Italic} text is internal. The \textbf{Length} is computed by the average number of tokens. \textbf{Multi} means the percentage of multi answer spans. \textbf{A-tok} mean the percentage of answer tokens.}\label{tab:data}
\begin{tabular}{lccccccc}
\hline
\textbf{Data}  & \textbf{Train} & \textbf{Test} & \textbf{Length} & \textbf{Field} & \textbf{Multi(\%)} & \textbf{A-tok(\%)} & \textbf{Lang}\\ \hline
\textit{License}  & 328 & 104 & 606 &  12 & 10.72 & 1.34 & JP  \\
Bidding \cite{Nguyen-IE-IJCNN-21}  & 82 & 20 & 1346 & 19 & 27.25 & 0.62 & JP  \\
CUAD-small \cite{Hendrycks-CUAD-21}  & 167 & 55 & 2678 & 17 & 6.86 & 0.65 & EN \\
\hline
\end{tabular}\vspace{-0.3cm}
\end{table}

\subsubsection{Public data}
We used two public business datasets for evaluation. \textbf{(i) Bidding} was created to simulate business scenarios in which a small number of training samples was sent from clients \cite{Nguyen-IE-IJCNN-21}. It includes 124 public Japanese\footnote{http://www.jogmec.go.jp/news/bid/search.php} bidding documents, in which 82 samples for training, 20 samples for validation, and 20 samples for testing. The number of targeted information/fields is 19. \textbf{(iii) CUAD-small} is a part of the CUAD dataset \cite{Hendrycks-CUAD-21}, that includes long English legal documents. Due to the computation resources, long documents with over 5000 words and questions that are less than 30 or more than 200 answers were excluded. Finally, the small set includes 167 training documents (originally 408 training documents), 55 for testing (originally 102 testing documents), and 17 fields (originally 41 fields).



\subsection{Baselines}
\subsubsection{Span extraction}
We compared our model to strong baselines of span extraction. We implemented three span-based IE methods formulated as MRC based on \textbf{BERT} \cite{DCLT-NAACL-19}, \textbf{DistilBERT} \cite{Sanh-DistillBERT-19}, and \textbf{ALBERT} \cite{Lan-ALBERT-ICLR-20}. These methods take each field-document pair as an input for training. We kept the same output vector of these models as 768 and used one MLP layer with the size of 128 for prediction. We also implemented six extensions based on PreLMs by adding CNN and BiLSTM on the top of the PreLMs. The output vector of PreLMs is 768. The kernel size of CNN is 3. The BiLSTM hidden size is 786. The classification uses a single MLP with the size of 128. The BERT+CNN model is similar to \cite{NLL-EAAI-21}.

We re-run two recent IE methods for business documents. \textbf{BERT-IE} \cite{Nguyen-IE-IJCNN-21} uses BERT for encoding, combined with a customized CNN layer for learning the local context in long documents and \textbf{Recur-spanIE} \cite{Nguyen-Recur-Span-DIL-21}  uses a recursive decoding method for extracting multiple spans. We kept the same setting of the original papers \cite{Nguyen-IE-IJCNN-21,Nguyen-Recur-Span-DIL-21}.


\subsubsection{Sequence labeling}
We implemented three strong sequence labeling methods for IE. \textbf{BERT-NER} \cite{Zhang-rapid-adaptation-BERT-20} uses BERT for encoding input documents. The training process was formulated as NER. \textbf{Flair+LSTM+CRF} \cite{Akbik-Flair-COLING-18} uses Flair for encoding, LSTM for learning hidden representation, and CRF for classification. We also tested \textbf{Recur-spanIE} \cite{Nguyen-Recur-Span-DIL-21} with the sequence labeling setting.

\subsection{Settings and evaluation metrics}
\subsubsection{Settings}

For License and Bidding, we used pre-trained Japanese BERT, Albert, and multilingual DistilBert. The corresponding pre-trained English base models were used for CUAD(s). For our model, we used  BERT (Japanese or English depending on the dataset) for encoding input documents. For span-IE, all query vectors were randomly initialized. We set the input sequence length as 384 and the stride is 128. The size of the output vector from PreLMs is 768. All models were optimized in 20 epochs by using the Adam method with a batch size of 16, the learning rate of $5e-5$. Our experiments were done by using a Tesla T4 GPU.

\subsubsection{Evaluation metrics}
We used two metrics for evaluation:
\begin{itemize}
    \item For span extraction, we followed the SQuAD format\footnote{https://rajpurkar.github.io/SQuAD-explorer/} to compute the F-score of fields. Note that since the SQuAD format does not work with multi-span questions, only the first span of each question will be considered.
    \item For sequence labeling, we followed the CoNLL definition to compute the F-score of each field (class). The final F-score is the micro average of F-scores over all fields.
\end{itemize}

\section{Results and Discussion}
This section reports the F-score comparison of our model and the baselines, and then shows the training and inference time. It also reports the coverage of multiple span extraction. It finally makes discussion on the outputs of IE models .

\subsection{F-score comparison}
In this section, we report the F-score comparison in two settings: \textit{span extraction} and \textit{sequence labeling}. The results are shown in Table \ref{tab:results}, in which SpanIE+NER is our model described in Fig. \ref{fig:model}, and SpanIE is that model but without the sequence labeling part (the part on the right).

\subsubsection{Span extraction}
The upper part of Table \ref{tab:results} shows the comparison of the span extraction. Our methods is better than the baselines in almost all cases, in which SpanIE+NER is the best on average. The improvement comes from two reasons. First, the method encodes short input fields and long documents in two channels, which can retain the meaning of short fields. Second, the combination of span extraction and sequence labeling also benefits the model. Compared to SpanIE, SpanIE+NER consistently achieves a higher F1-score of 0.05-0.1, at the cost of only a simple token classification head. It shows the NER part helps to improve the performance of the SpanIE part for predicting the first answer spans by contributing better gradient during training.

\begin{table}[!h]
\centering
\setlength{\tabcolsep}{1.9pt}
\caption{The F-scores comparison of our model and baselines. Bid(dev) and Bid(test) are the development and test sets of the bidding dataset. CUAD(s) means the small version. \textbf{Bold} text is the best. \underline{Underline} text is the second best. $\dagger$ means that our method is significantly better with $p \leq 0.05$ by using the pair-wise $t$-test.}\label{tab:results}
\begin{tabular}{lccccc}
\hline
\textbf{Method} & \textbf{License} & \textbf{Bid(dev)} & \textbf{Bid(test)}  & \textbf{CUAD(s)} & \textbf{Average} \\ \hline

\multicolumn{6}{c}{\textbf{Span extraction}} \\ \hline
BERT-QA                             & 0.7849 & 0.8484 & 0.8915  & 0.8104 & 0.8632$^{\dagger}$ \\
DistilBERT                          & 0.8120 & 0.8130 & 0.8366  & 0.8369 & 0.8553$^{\dagger}$ \\ 
ALBERT                              & 0.8276 & 0.8509 & 0.8658  & 0.8344 & 0.8718$^{\dagger}$ \\ 

BERT+CNN \cite{NLL-EAAI-21}  & 0.8062 & 0.8529 & 0.8853   & 0.7993 & 0.8646$^{\dagger}$ \\
DistilBERT+CNN                      & 0.8073 & 0.7550 & 0.8433   & \textbf{0.8491} & 0.8468$^{\dagger}$ \\
ALBERT+CNN                          & 0.8092 & 0.7497 & 0.7900   & 0.8336 & 0.8329$^{\dagger}$\\ 

BERT+BiLSTM                         & 0.7946 & 0.8569 & 0.8945 & 0.8380 & 0.8733$^{\dagger}$ \\
DistilBERT+BiLSTM                   & 0.8190 & 0.8362 & 0.8720 & 0.8385 & 0.8695$^{\dagger}$ \\
ALBERT+BiLSTM                       & 0.8897 & 0.7672 & 0.8616 & 0.8204 & 0.8640$^{\dagger}$ \\
BERT-IE \cite{Nguyen-IE-IJCNN-21}   & 0.8066 & 0.8344 & 0.8969 & 0.8073 & 0.8655$^{\dagger}$ \\
Recur-spanIE \cite{Nguyen-Recur-Span-DIL-21}
                                    & 0.9345 & \textbf{0.8990} & 0.9049 & \underline{0.8466} & 0.9107 \\ \hline

SpanIE                              & \underline{0.9353} & 0.8872 & \underline{0.9191} & 0.8360 & \underline{0.9123} \\ 
SpanIE+NER                          & \textbf{0.9462} & \underline{0.8974} & \textbf{0.9244} &  0.8417 & \textbf{0.9189} \\  \hline \hline

\multicolumn{6}{c}{\textbf{Sequence labeling (NER)}} \\ \hline
BERT-NER \cite{Zhang-rapid-adaptation-BERT-20}  & \underline{0.8994} & 0.6704 & 0.7043 & 0.5112 & 0.7459$^{\dagger}$ \\  
Flair+LSTM+CRF \cite{Akbik-Flair-COLING-18}     & 0.8757 &  \underline{0.7601} & 0.6965 & 0.3291 & 0.7153$^{\dagger}$ \\
Recur-spanIE \cite{Nguyen-Recur-Span-DIL-21}    & 0.8493 &  0.7587 & \underline{0.8302} & \underline{0.5347} & \underline{0.7886}$^{\dagger}$ \\  \hline
SpanIE                                          & 0.7950 & 0.6666 & 0.7427 & 0.4566 & 0.6706$^{\dagger}$ \\ 
SpanIE+NER                                      & \textbf{0.9051}  & \textbf{0.7760} & \textbf{0.8394} & \textbf{0.5768} & \textbf{0.8091} \\  \hline

\end{tabular}
\end{table}

The span IE baselines also achieve quite promising results. Recur-spanIE \cite{Nguyen-Recur-Span-DIL-21} is the second-best on the average. This is because this model uses a recursive decoding method that can extract multiple spans. However, its performance is still lower than our model. DistilBERT+CNN is the best on CUAD (small) because it is empowered by the combination of DistilBERT and CNN which is potential for IE of business documents \cite{NLL-EAAI-21,Nguyen-IE-IJCNN-21}.
The original PreLMs also obtain competitive results. It again confirms the advantage of PreLMs for the IE task \cite{DCLT-NAACL-19,Sanh-DistillBERT-19,Lan-ALBERT-ICLR-20}. The combination of PreLMs with CNN and BiLSTM does not show the best results. A possible reason is that the knowledge from PreLMs is enough and adding additional layers seems to be saturated.

\subsubsection{Sequence labeling}
We next compared our model to strong baselines of NER. Results in the bellow part of Table \ref{tab:results} show that our model is the best on the average of F-scores. This again validates our assumption. Being only able to extract single spans, the SpanIE model performs poorly on the multi-span extraction setting, which is very common in NER data sets. Here, its performance is included solely for the sake of completeness. The recursive span IE method \cite{Nguyen-Recur-Span-DIL-21} is still the second best, demonstrating the potential of span extraction approach for IE tasks. As our expectation, common NER methods, e.g. Flair+LSTM+CRF \cite{Akbik-Flair-COLING-18} is not the best because it can not handle long business documents with sparse information. Note that Flair+LSTM+CRF \cite{Akbik-Flair-COLING-18} is a strong method with the position of 18 over 60 teams of CoNLL 2003.\footnote{The results were derived from: https://paperswithcode.com/sota/named-entity-recognition-ner-on-conll-2003. Accessed on October 7th, 2021.} Also, we can not re-run the best NER model named ACE+document-context \cite{Wang-ACE-NER-ACL-2021} due to computing resources. The SpanIE+NER showed the best result for the most cases and those results are significantly higher than both BERT-NER and SpandIE, which describes that the NER part supports the SpanIE part on handling the multi-span extraction problem. The better score of SpanIE+NER compared to BERT-NER come from the high precision of the SpanIE part.

On a further note, as our SpanIE+NER model was not tailored to named entity recognition, we do not aim to achieve SOTA results on NER benchmark data sets. Instead, these experiments compare our SpanIE+NER to popular NER models on information extraction for business documents (which are usually long and contains short and sparse entities). Our SpanIE+NER model without the SpanIE part would become BERT-NER, which under-performs our method. This shows that in SpanIE+NER, the SpanIE part improves the performance of the NER part in sequence labeling setting. Our SpanIE model - which is the SpanIE+NER without the NER part - is not suitable for sequence labeling, but is still included for the sake of completeness.



\subsection{Training and inference time}
\subsubsection{Training and inference}
As mentioned, our uniqueness is to speed up the training and inference of span extraction. To show this aspect, we observed the training and inference time of four models: BERT-QA \cite{DCLT-NAACL-19}, BERT-IE cite{Nguyen-IE-IJCNN-21}, our SpanIE, and our SpanIE+NER. Table \ref{tab:training-inference} shows the training and inference time of normal span IE (BERT-QA and BERT-IE) and our span IE methods.\footnote{We did not include the comparison of NER methods because they are quite fast.} 
\begin{table}[!h]
\centering
\setlength{\tabcolsep}{3.8pt}
\caption{\label{tab:training-inference} Train epoch time and inference time. The format is \texttt{hh:mm:ss}.}
\begin{tabular}{ccccccc}
\hline
 & \textbf{Method}  & \textbf{License} & \textbf{Bidding(d)} & \textbf{Bidding(t)} & \textbf{CUAD(s)}\\ \hline
\multirow{4}{*}{\begin{tabular}[c]{@{}l@{}}Epoch\\ time\end{tabular}} 
    & BERT-QA \cite{DCLT-NAACL-19}  & 0:11:50  & 0:42:38 & 0:44:35  & 1:43:59 \\
    & BERT-IE \cite{Nguyen-IE-IJCNN-21}  & 0:09:59  & 0:44:29  & 0:44:35 & 1:55:56 \\
    & SpanIE     & \underline{0:00:56} & \textbf{0:03:19} & \underline{0:03:30} & \textbf{0:05:53} \\
   & SpanIE+NER   & \textbf{0:00:54} & \underline{0:03:26}  & \textbf{0:03:26} & \underline{0:06:12}  \\
\hline
\multirow{4}{*}{\begin{tabular}[c]{@{}l@{}}Infer\\ time\end{tabular}} 
    & BERT-QA \cite{DCLT-NAACL-19}   & 0:01:45  & 0:02:55  & 0:03:39 &	0:10:33 \\
    & BERT-IE \cite{Nguyen-IE-IJCNN-21}   & 0:01:19 & 0:03:02  & 0:02:50 & 0:13:06 \\
    & SpanIE   & \textbf{0:00:10}  & \textbf{0:00:14}  & \textbf{0:00:13}  & \underline{0:01:01} \\ 
   & SpanIE+NER   & \textbf{0:00:10} & \textbf{0:00:14}  & \underline{0:00:14}  & \textbf{0:00:55} \\ \hline
\end{tabular}

\end{table}

Our span IE methods are much faster than the normal span methods in both training and inference. For each epoch, SpanIE methods take a very short time to complete while the normal span IE methods require a much longer time. The inference shares the same trend. It comes from the separation encoding of short fields and documents. There is no significant difference between SpanIE and SpanIE+NER because they share the same separated encoding method and the NER part does not significantly increase the number of parameters.

\subsubsection{Convergence time}
We also observed the convergence time of span-based IE methods. Due to no significant difference between BERT-QA \cite{DCLT-NAACL-19} and BERT-IE \cite{Nguyen-IE-IJCNN-21}; and SpanIE and SpanIE+NER, we only report BERT-QA and SpanIE+NER.

\begin{figure*}[!h]
    \centering
    \subfigure[License]{\label{fig:license}\includegraphics[width=56mm]{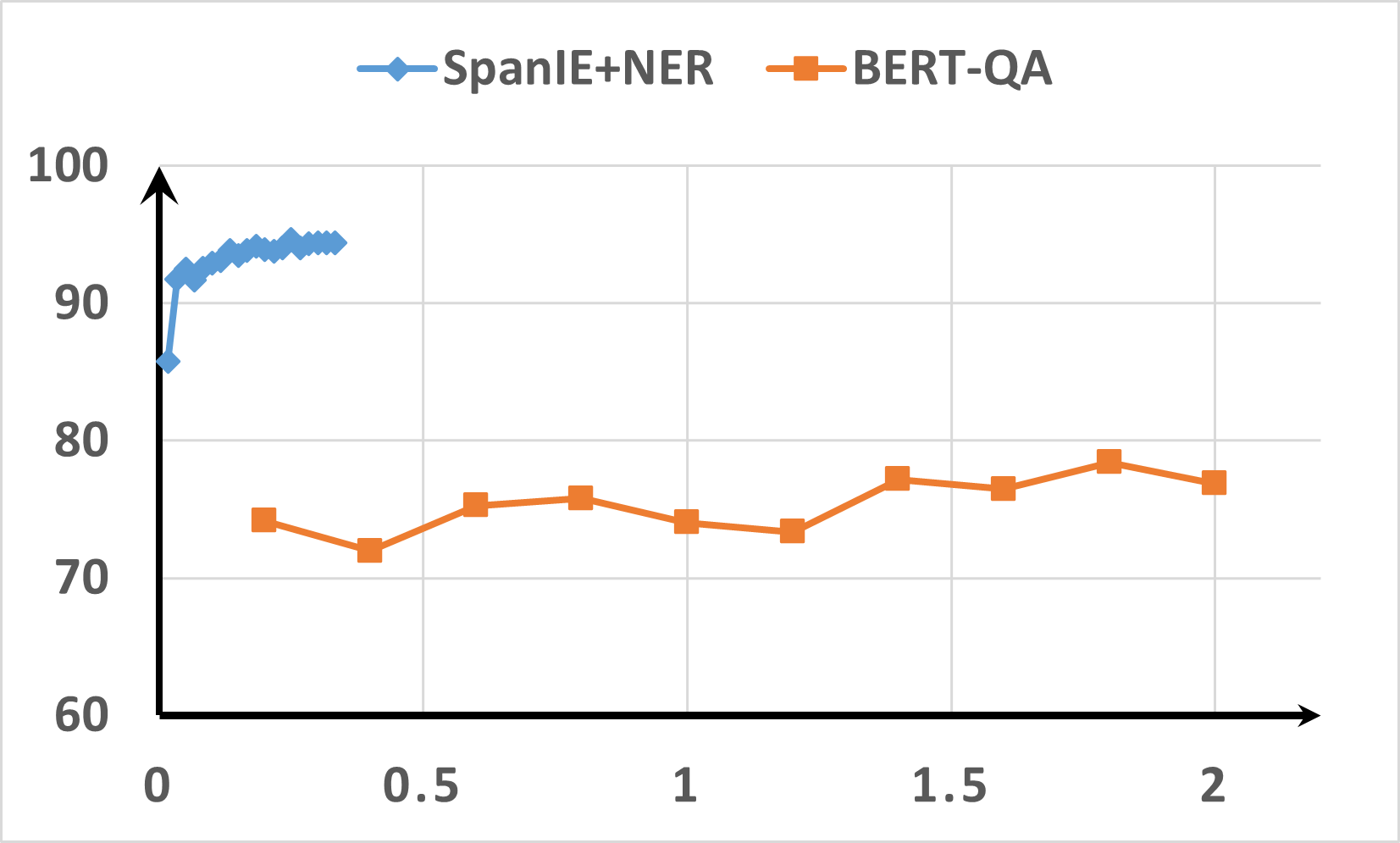}}
    \subfigure[Bidding (train)]{\label{fig:cin_dev}\includegraphics[width=56mm]{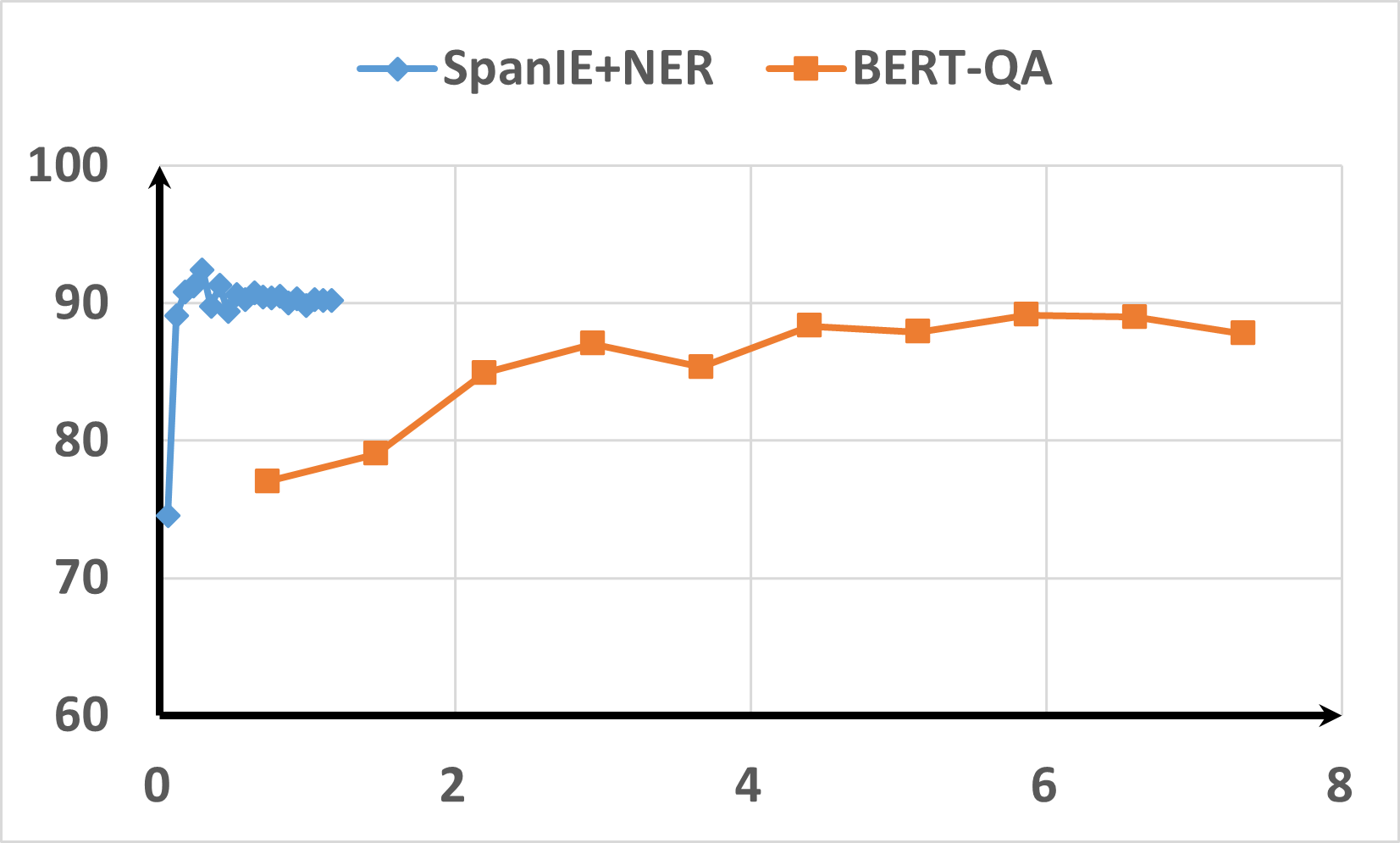}}
    \subfigure[CUAD-small]{\label{fig:cuad}\includegraphics[width=56mm]{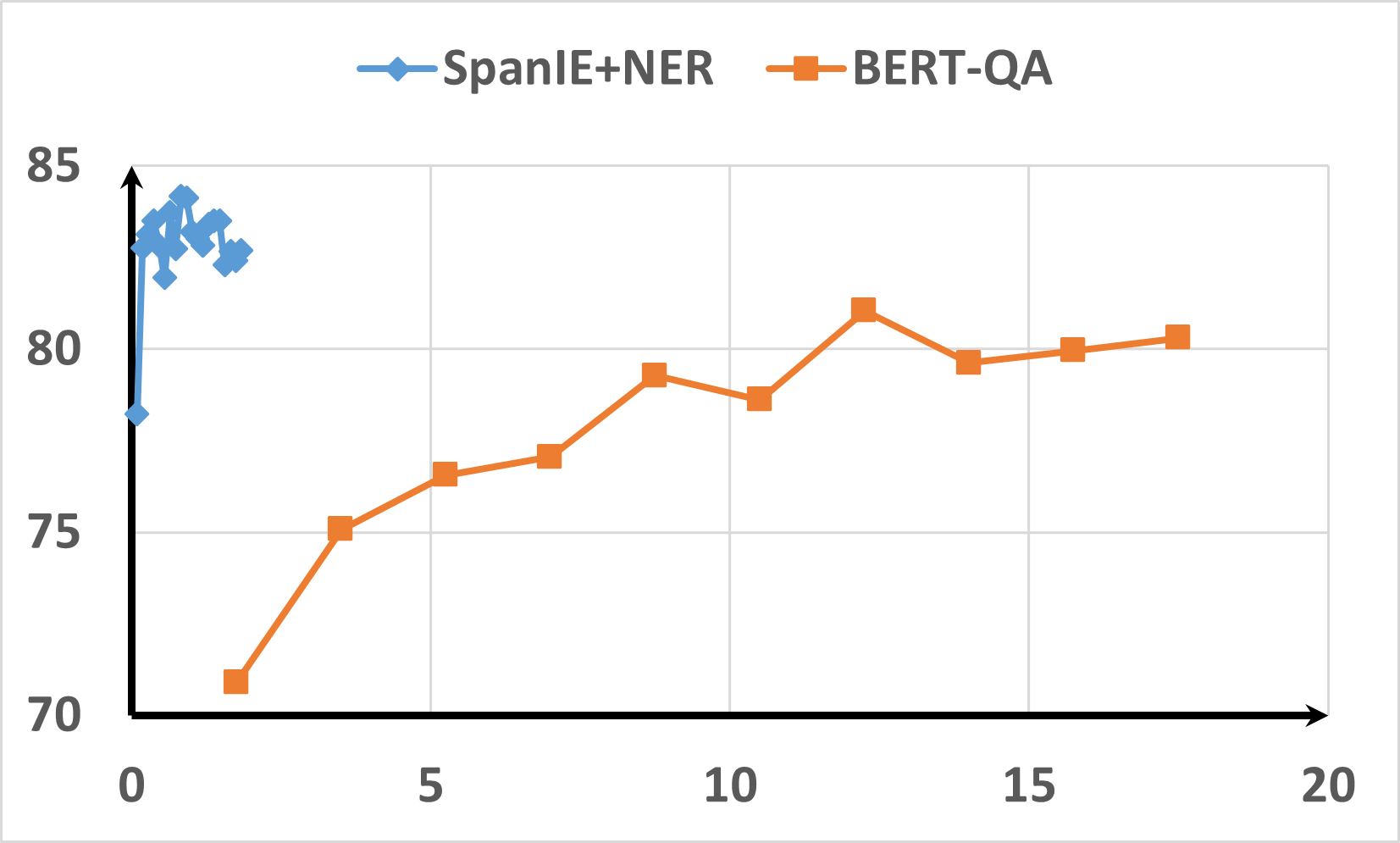}}
        \caption{The convergence of the normal span IE using BERT-QA and SpanIE+NER. The $x$-axis is the time in hours; the $y$-axis is the F-score.}\label{fig:convergence-time}\vspace{-0.2cm}
\end{figure*}

The trend in Figure \ref{fig:convergence-time} supports the results in Table \ref{tab:training-inference}, in which the SpanIE+NER model requires a shorter time for convergence than the normal span IE method (BERT-QA). The reason is that the separation encoding and multiple general attentions help the model to converge quickly. Also, the sequence labeling part supports the span IE part for effective learning hidden patterns of multiple spans.

\subsection{Multiple span extraction coverage}
We observed the ability of our model to extract multiple answer spans. To do that, we computed the recall scores. Table \ref{tab:coverage} shows that the SpanIE+NER and Recur-spanIE methods achieves the highest scores.
\begin{table}[!h]
\centering
\setlength{\tabcolsep}{4pt}
\caption{The recall scores of multiple span extraction from IE models. \textbf{Bold} text is the best. \underline{Underline} text is the second best.}\label{tab:coverage}
\begin{tabular}{lccccc}\hline
\textbf{Data}   & \textbf{Flair} & \textbf{Recur-spanIE} & \textbf{Span-IE} & \textbf{NER} & \textbf{Span-IE+NER} \\ \hline
License  & \textbf{0.8555}   & 0.7452     & 0.6513    & 0.8333    & \underline{0.8525}     \\
Bidding   & 0.6308   & \textbf{0.8594}     & 0.6712    & 0.6236    & \textbf{0.8594}     \\
CUAD-s    & 0.2390   & \underline{0.3176}     & 0.2660    & 0.2760    & \textbf{0.3708}     \\ \hline
\end{tabular}
\end{table}
This is because these methods consider multiple span extraction. It supports the results in Table \ref{tab:results} in which SpanIE+NER and Recur-spanIE obtain promising results. Flair+LSTM+CRF and NER (BERT-NER) are also good for dealing with multiple spans because it was designed for sequence labeling. The SpanIE method outputs poor results because it was designed to only extract single spans.

\subsection{Output observation}
Table \ref{tab:output} shows the output of SpanIE and SpanIE+NER to extract the information of the fields \texttt{\begin{CJK}{UTF8}{min}施設名\end{CJK}} and \texttt{\begin{CJK}{UTF8}{min}質問箇所TEL/FAX\end{CJK}} from a bidding document.

\begin{table}[!h]
\centering
\setlength{\tabcolsep}{4pt}
\caption{The output of SpanIE and SpanIE+NER. Correct answer spans are highlighted in \textcolor{red}{red color}. \texttt{xxx} are hidden information due to confidential reasons.}\label{tab:output}
\begin{tabular}{lp{6.3cm}} \hline
\textbf{Field}          & \begin{CJK}{UTF8}{min} 施設名 \end{CJK} (Question about Name of facility ). \\             \hline
\multirow{8}{*}{\begin{tabular}[c]{@{}l@{}}\textbf{Input text}\end{tabular}}   & \begin{CJK}{UTF8}{min} （４）需要場所 北海道苫小牧市字静川３０８番地
\textcolor{red}{苫小牧東部国家石油備蓄基地} （５ ） 入札方法 入札金額は、各社において設定する契約電力に対する単一の単価（ｋＷ単
価）及び使用電力量に対する単価 (\textbf{Translation to English:} (4) Demand location 308 Shizukawa, Tomakomai-shi, Hokkaido
\textcolor{red}{Tomakomai Eastern National Petroleum Stockpiling Base}
(5) Bid method The bid amount is a single unit price (kW unit price) for the contracted power set by each company and the unit price for the amount of power used. 
\end{CJK} 
\\ \hline
                     
\textbf{SpanIE}  & \begin{CJK}{UTF8}{min} 苫小牧東部国家石油備蓄基地 \end{CJK}  \\ \hline
\textbf{SpanIE+NER}  &   \begin{CJK}{UTF8}{min} 苫小牧東部国家石油備蓄基地   \end{CJK} \\  \hline \hline

\textbf{Field}          & \begin{CJK}{UTF8}{min} 質問箇所TEL/FAX \end{CJK} (Question of TEL/FAX). \\             \hline
\multirow{8}{*}{\begin{tabular}[c]{@{}l@{}}\textbf{Input text}\end{tabular}}   & \begin{CJK}{UTF8}{min} 〒105-0001東京都港区虎ノ門二丁目。独立行政法人石油天然ガス。 金属鉱物資源機構。石油備蓄部基地管理課。\textcolor{red}{(電 話)03-xxxx-8517}。 \textcolor{red}{(FAX)03-xxxx-8064}。(Email)xxxx@jogmec.go.jp。(\textbf{Translation to English:} 2-Chome, Toranomon, Minato-ku, Tokyo 105-0001. Japan Oil, Gas and Metals National Corporation., Base Management Division, Oil Storage Department. \textcolor{red}{TEL: 03-xxxx-8517}. \textcolor{red}{(FAX)03-xxxx-8064}. (Email)xxxx@jogmec.go.jp.)
\end{CJK} 
\\ \hline
                     
\textbf{SpanIE}  & \begin{CJK}{UTF8}{min} (電話)03-xxxx-8517 \end{CJK} (TEL: 03-xxxx-8517).  \\ \hline
\multirow{2}{*}{\begin{tabular}[c]{@{}l@{}}\textbf{SpanIE+NER}\end{tabular}}  &   \begin{CJK}{UTF8}{min} (電話)03-xxxx-8517   \end{CJK} (TEL: 03-xxxx-8517). \\
            & \begin{CJK}{UTF8}{min} (FAX)03-xxxx-8064.  \end{CJK}.  \\ \hline

\end{tabular}\vspace{-0.5cm}
\end{table}

The upper part of Table \ref{tab:output} shows the outputs of SpanIE and SpanIE+NER. The two models share the same output. This is because the example contains only one answer for one question. As the result, the two models can correctly extract the span. On the one hand, in the below part, the SpanIE method can only extract the first answer span (for both examples) because it was formulated as a span extraction (MRC) task. Hence, the second span was ignored. In contrast, the SpanIE+NER can extract correctly two answer spans, which is described in the prediction corresponding to the first example. This is because it combines advantage of span extraction and sequence labeling. The span extraction part correctly extracts the first span, while the sequence labeling can extract the second span by using the aggregation algorithm. The joint learning process also allows the model to learn hidden patterns of answer spans based on the input query field.




\section{Conclusion}
This paper introduces a new query model for IE from business documents. The model can deal with multiple span extraction and speed up the training and inference process. Promising results on four business datasets in English and Japanese show two important points. First, the combination of span extraction and sequence labeling benefit multiple span extraction. Second, the separation encoding of query fields and documents speeds up the model. The method can be applied to most existing span extraction models for IE. Future work will investigate decoding methods for improving the quality of multiple span extraction.

\section*{Acknowledgment}

This research is funded by Hung Yen University of Technology and Education under the grant number UTEHY.L.2022.02.


\bibliographystyle{IEEEtran}
\bibliography{ie-ref}

\begin{thebibliography}{10}
\providecommand{\url}[1]{#1}
\csname url@samestyle\endcsname
\providecommand{\newblock}{\relax}
\providecommand{\bibinfo}[2]{#2}
\providecommand{\BIBentrySTDinterwordspacing}{\spaceskip=0pt\relax}
\providecommand{\BIBentryALTinterwordstretchfactor}{4}
\providecommand{\BIBentryALTinterwordspacing}{\spaceskip=\fontdimen2\font plus
\BIBentryALTinterwordstretchfactor\fontdimen3\font minus
  \fontdimen4\font\relax}
\providecommand{\BIBforeignlanguage}[2]{{%
\expandafter\ifx\csname l@#1\endcsname\relax
\typeout{** WARNING: IEEEtran.bst: No hyphenation pattern has been}%
\typeout{** loaded for the language `#1'. Using the pattern for}%
\typeout{** the default language instead.}%
\else
\language=\csname l@#1\endcsname
\fi
#2}}
\providecommand{\BIBdecl}{\relax}
\BIBdecl

\bibitem{APM-ACL-IJCNLP-15}
G.~Angeli, M.~J. Premkumar, and C.~D. Manning, ``Leveraging linguistic
  structure for open domain information extraction,'' in \emph{Proceedings of
  the 53rd Annual Meeting of the Association for Computational Linguistics and
  the 7th International Joint Conference on Natural Language Processing, pp.
  344-354}, 2015.

\bibitem{LBSKD-NER-NACCAL-16}
G.~Lample, M.~Ballesteros, S.~Subramanian, K.~Kawakami, and C.~Dyer, ``Neural
  architectures for named entity recognition,'' in \emph{Proceedings of the
  2016 Conference of the North American Chapter of the Association for
  Computational Linguistics: Human Language Technologies, pp. 260-270}, 2016.

\bibitem{Ram-Few-shot-span-IE-21}
O.~Ram, Y.~Kirstain, J.~Berant, A.~Globerson, and O.~Levy, ``Few-shot question
  answering by pretraining span selection,'' in \emph{arXiv preprint
  arXiv:2101.00438}, 2021.

\bibitem{Nguyen-Recur-Span-DIL-21}
T.-A.~D. Nguyen, H.~M. Vu, N.~H. Son, and M.-T. Nguyen, ``A span extraction
  approach for information extraction on visually-rich documents,'' in
  \emph{Document Analysis and Recognition - ICDAR 2021 Workshops, pp. 353-363},
  2021.

\bibitem{Li-Span-NER-21}
F.~Li, Z.~Lin, M.~Zhang, and D.~Ji, ``A span-based model for joint overlapped
  and discontinuous named entity recognition,'' in \emph{Proceedings of the
  59th Annual Meeting of the Association for Computational Linguistics and the
  11th International Joint Conference on Natural Language Processing, pp.
  4814–4828}, 2021.

\bibitem{Zhang-rapid-adaptation-BERT-20}
R.~Zhang, W.~Yang, L.~Lin, Z.~Tu, Y.~Xie, Z.~Fu, Y.~Xie, L.~Tan, K.~Xiong, and
  J.~Lin, ``Rapid adaptation of bert for information extraction on
  domain-specific business documents,'' in \emph{arXiv preprint
  arXiv:2002.01861}, 2020.

\bibitem{Karamanolakis-TXtract-ACL-20}
G.~Karamanolakis and X.~L.~D. Jun~Ma, ``Txtract: Taxonomy-aware knowledge
  extraction for thousands of product categories,'' in \emph{Proceedings of the
  58th Annual Meeting of the Association for Computational Linguistics, pp.
  8489–8502}, 2020.

\bibitem{DCLT-NAACL-19}
J.~Devlin, M.-W. Chang, K.~Lee, and K.~Toutanova, ``Bert: Pre-training of deep
  bidirectional transformers for language understanding,'' in \emph{Proceedings
  of the 2019 Conference of the North American Chapter of the Association for
  Computational Linguistics: Human Language Technologies, Volume 1 (Long and
  Short Papers), pp. 4171-4186}, 2019.

\bibitem{Xu-LayoutLM-20}
Y.~Xu, M.~Li, L.~Cui, S.~Huang, F.~Wei, and M.~Zhou, ``Layoutlm: Pre-training
  of text and layout for document image understanding,'' in \emph{Proceedings
  of the 26th ACM SIGKDD International Conference on Knowledge Discovery \&
  Data Mining, pp. 1192-1200}, 2020.

\bibitem{Herbert-dig-tran-17}
L.~Herbert, ``Digital transformation: Build your organization's future for the
  innovation age,'' Bloomsbury Publishing, Tech. Rep., 2017.

\bibitem{NLL-EAAI-21}
M.-T. Nguyen, D.~T. Le, and L.~Le, ``Transformers-based information extraction
  with limited data for domain-specific business documents,'' \emph{Engineering
  Applications of Artificial Intelligence, 97, 104100}, 2021.

\bibitem{SKM-CONLL-03}
E.~Sang, T.~Kim, and F.~D. Meulder, ``Introduction to the conll-2003 shared
  task: Language-independent named entity recognition,'' in \emph{Proceedings
  of the Seventh Conference on Natural Language Learning at HLT-NAACL 2003},
  2003.

\bibitem{Nguyen-IE-IJCNN-21}
M.-T. Nguyen, L.~T. Linh, D.~T. Le, N.~H. Son, D.~H.~T. Duong, B.~C. Minh, and
  A.~Shojiguchi, ``Information extraction of domain-specific business documents
  with limited data,'' in \emph{2021 International Joint Conference on Neural
  Networks (IJCNN), pp. 1-8. IEEE}, 2021.

\bibitem{Li-MRC-NER-ACL-20}
X.~Li, J.~Feng, Y.~Meng, Q.~Han, F.~Wu, and J.~Li, ``A unified mrc framework
  for named entity recognition,'' in \emph{Proceedings of the 58th Annual
  Meeting of the Association for Computational Linguistics, pp. 5849-5859},
  2020.

\bibitem{Huang-BiLSTM-CRF-NER-15}
Z.~Huang, W.~Xu, and K.~Yu, ``Bidirectional lstm-crf models for sequence
  tagging,'' in \emph{arXiv preprint arXiv:1508.01991}, 2015.

\bibitem{Katiyar-Nested-NER-NAACL-18}
A.~Katiyar and C.~Cardie, ``Nested named entity recognition revisited,'' in
  \emph{Proceedings of the 2018 Conference of the North American Chapter of the
  Association for Computational Linguistics: Human Language Technologies,
  Volume 1 (Long Papers), pp. 861–871}, 2018.

\bibitem{Liu-NER-AAAI-18}
L.~Liu, J.~Shang, X.~Ren, F.~Xu, H.~Gui, J.~Peng, and J.~Han, ``Empower
  sequence labeling with task-aware neural language model,'' in
  \emph{Proceedings of the AAAI Conference on Artificial Intelligence, vol. 32,
  no. 1}, 2018.

\bibitem{JWA-NAACL-18}
M.~Ju, M.~Miwa, and S.~Ananiadou, ``A neural layered model for nested named
  entity recognition,'' in \emph{Proceedings of the 2018 Conference of the
  North American Chapter of the Association for Computational Linguistics:
  Human Language Technologies, Volume 1 (Long Papers), pp. 1446-1459}, 2018.

\bibitem{Collobert-NLP-JML-11}
R.~Collobert, J.~Weston, L.~Bottou, M.~Karlen, K.~Kavukcuoglu, and P.~Kuksa,
  ``Natural language processing (almost) from scratch,'' \emph{Journal of
  machine learning research 12, no. ARTICLE: 2493-2537}, 2011.

\bibitem{Akbik-Flair-COLING-18}
A.~Akbik, D.~Blythe, and R.~Vollgraf, ``Contextual string embeddings for
  sequence labeling,'' in \emph{Proceedings of the 27th international
  conference on computational linguistics, pp. 1638-1649}, 2018.

\bibitem{btg1023}
J.-D. Kim, T.~Ohta, Y.~Tateisi, and J.~Tsujii, ``{GENIA corpus—a semantically
  annotated corpus for bio-textmining},'' \emph{Bioinformatics}, vol.~19, 07
  2003.

\bibitem{Zheng-Boundary-Nested-NER-EMNLP-19}
C.~Zheng, Y.~Cai, J.~Xu, H.~fung Leung, and G.~Xu, ``A boundary-aware neural
  model for nested named entity recognition,'' in \emph{Proceedings of the 2019
  Conference on Empirical Methods in Natural Language Processing and the 9th
  International Joint Conference on Natural Language Processing, pp. 357-366},
  2019.

\bibitem{Fu-Nested-NER-TreeCRF-AAAI-21}
Y.~Fu, C.~Tan, M.~Chen, S.~Huang, and F.~Huang, ``Nested named entity
  recognition with partially-observed treecrfs,'' in \emph{Proceedings of the
  AAAI Conference on Artificial Intelligence, vol. 35, no. 14, pp.
  12839-12847}, 2021.

\bibitem{Rajpurkar-SQuAD-EMNLP-16}
P.~Rajpurkar, J.~Zhang, K.~Lopyrev, and P.~Liang, ``Squad: 100,000+ questions
  for machine comprehension of text,'' in \emph{Proceedings of the 2016
  Conference on Empirical Methods in Natural Language Processing, pp.
  2383–2392}, 2016.

\bibitem{aly-etal-2021-leveraging}
\BIBentryALTinterwordspacing
R.~Aly, A.~Vlachos, and R.~McDonald, ``Leveraging type descriptions for
  zero-shot named entity recognition and classification,'' in \emph{Proceedings
  of the 59th Annual Meeting of the Association for Computational Linguistics
  and the 11th International Joint Conference on Natural Language Processing
  (Volume 1: Long Papers)}.\hskip 1em plus 0.5em minus 0.4em\relax Online:
  Association for Computational Linguistics, Aug. 2021, pp. 1516--1528.
  [Online]. Available: \url{https://aclanthology.org/2021.acl-long.120}
\BIBentrySTDinterwordspacing

\bibitem{Luong-Att-EMNLP-15}
M.-T. Luong, H.~Pham, and C.~D. Manning, ``Effective approaches to
  attention-based neural machine translation,'' in \emph{Proceedings of the
  2015 Conference on Empirical Methods in Natural Language Processing, pp.
  1412-1421}, 2015.

\bibitem{Hendrycks-CUAD-21}
D.~Hendrycks, C.~Burns, A.~Chen, and S.~Ball, ``Cuad: An expert-annotated nlp
  dataset for legal contract review,'' in \emph{arXiv:2103.06268}, 2021.

\bibitem{Sanh-DistillBERT-19}
V.~Sanh, L.~Debut, J.~Chaumond, and T.~Wolf, ``Distilbert, a distilled version
  of bert: Smaller, faster, cheaper and lighter,'' in \emph{arXiv preprint
  arXiv:1910.01108}, 2019.

\bibitem{Lan-ALBERT-ICLR-20}
Z.~Lan, M.~Chen, S.~Goodman, K.~Gimpel, P.~Sharma, and R.~Soricut, ``Albert: A
  lite bert for self-supervised learning of language representations,'' in
  \emph{Proceedings of The International Conference on Learning
  Representations}, 2020.

\bibitem{Wang-ACE-NER-ACL-2021}
X.~Wang, Y.~Jiang, N.~Bach, T.~Wang, Z.~Huang, F.~Huang, and K.~Tu, ``Automated
  concatenation of embeddings for structured prediction,'' in \emph{Proceedings
  of the 59th Annual Meeting of the ACL and the 11th International Joint
  Conference on Natural Language Processing, pp. 2643–2660}, 2021.

\end{thebibliography}

\end{document}